# Physics-Informed Linear Model (PILM): Analytical Representations and Application to Crustal Strain Rate Estimation

Tomohisa Okazaki
RIKEN Center for Advanced Intelligence Project

**Abstract**

Many physical systems are described by partial differential equations (PDEs), and solving these equations and estimating their coefficients or boundary conditions (BCs) from observational data play a crucial role in understanding the associated phenomena. Recently, a machine learning approach known as physics-informed neural network, which solves PDEs using neural networks by minimizing the sum of residuals from the PDEs, BCs, and data, has gained significant attention in the scientific community. In this study, we investigate a physics-informed linear model (PILM) that uses linear combinations of basis functions to represent solutions, thereby enabling an analytical representation of optimal solutions. The PILM was formulated and verified for illustrative forward and inverse problems including cases with uncertain BCs. Furthermore, the PILM was applied to estimate crustal strain rates using geodetic data. Specifically, physical regularization that enforces elastic equilibrium on the velocity fields was compared with mathematical regularization that imposes smoothness constraints. From a Bayesian perspective, mathematical regularization exhibited superior performance. The PILM provides an analytically solvable framework applicable to linear forward and inverse problems, underdetermined systems, and physical regularization.

## 1. Introduction

Partial differential equations (PDEs) are ubiquitous in physical phenomena and are widely used to model various processes. In solid Earth geosciences, quantitative analyses based on PDEs have played a fundamental role in understanding the dynamics of Earth systems (e.g., Aki and Richards, 2002; Turcotte and Schubert, 2014). In recent years, inspired by the great success of deep learning in information technologies, machine learning approaches have attracted growing interest for solving PDEs in scientific and engineering contexts (Brunton and Kutz, 2024). Machine learning is expected to have the flexibility to assimilate observational data into physical models.

Suppose that an unknown solution $u(x)$ obeys a system of PDEs and boundary conditions (BCs) defined by

$$\mathcal{A}[u] = f(x) \quad (x \in \Omega), \tag{1a}$$

$$\mathcal{B}[u] = g(x) \quad (x \in \partial\Omega), \tag{1b}$$

where $\mathcal{A}$ and $\mathcal{B}$ are differential operators, $f$ and $g$ are known functions, and $\Omega$ and $\partial\Omega$ denote the domain and its boundary, respectively. Solving this system of equations can be reformulated as the following optimization problem:

$$u = \operatorname{argmin} \mathcal{L}[u], \tag{2a}$$

$$\mathcal{L}[u] = \mathcal{L}_{\text{PDE}}[u] + \mathcal{L}_{\text{BC}}[u] = \int_\Omega \left(\mathcal{A}[u] - f(x)\right)^2 dx + \int_{\partial\Omega} \left(\mathcal{B}[u] - g(x)\right)^2 dx. \tag{2b}$$

Here, $\mathcal{L}[u]$ is a loss functional that attains a global minimum of zero if and only if $u(x)$ is the solution to eq. (1). To address the infinite-dimensional functional space of $u$, machine learning approaches typically follow the following steps: (i) Represent the solution using parametric functions, $u(x; \mathbf{a})$, where $\mathbf{a} \in \mathbb{R}^M$ denotes the vector of model parameters; (ii) Represent the loss function in terms of the model parameters, $L(\mathbf{a}) = \mathcal{L}[u(x; \mathbf{a})]$; (iii) Identify the optimal model parameters by minimizing the loss function, $\mathbf{a}^* = \operatorname{argmin} L(\mathbf{a})$. When observational data are available, data assimilation and inversion analysis can be performed in a unified framework by adding the data residual term to $\mathcal{L}[u]$. This class of approaches is referred to as physics-informed machine learning (PIML) (Karniadakis et al., 2021).

A seminal implementation of PIML is the physics-informed neural network (PINN) (Raissi et al., 2019). By leveraging the flexible representation capacity of neural networks and the availability of advanced programming and computational environments, PINNs have emerged as powerful and versatile methods for solving nonlinear and stiff PDEs (Sharma et al., 2023). Automatic differentiation is a key technique that enables accurate and efficient computation of derivatives. However, the integrals in eq. (2b) are approximated by sums over discrete collocation points, and the minimization in eq. (2a) is performed using a stochastic gradient method (Table 1). The accuracy of the solutions depends strongly on their choice, and efficient and reliable techniques have been actively investigated (e.g., McClenny and Braga-Neto, 2023; Müller and Zeinhofer, 2023).

Many physical phenomena can be described by linear PDEs, to which linear machine learning models can offer mathematically sophisticated PIML approaches. One effective approach is the physics-informed Gaussian process (PIGP) (Raissi et al., 2017). By appropriately defining the kernel (covariance) functions of Gaussian processes based on linear operators, this method yields a closed-form optimal solution. Although the kernel functions can be derived analytically, collocation points must be sampled to calculate the posterior distributions (Table 1).

In this study, as another form of linear PIML, we explore a physics-informed linear model (PILM) that employs a fixed set of basis functions. With a suitable choice of basis functions, the derivatives and integrals in eq. (2b) can be calculated analytically, and the minimization problem in eq. (2a) admits a closed-form least-squares solution (Table 1). Thus, the optimal solution for the specified model can be derived analytically without approximation. Such solvability is valuable for analyzing both the theoretical and practical properties of the model and for ensuring the reliability of the results.

Section 2 presents the formulation of the PILM for both forward and inverse problems. Rather than adopting general abstract notation, we illustrate the method using a specific ordinary differential equation (ODE) (damped oscillation) and a PDE (diffusion equation) to demonstrate how analytical representations can be derived for these problems. Section 3 applies the PILM to estimate the strain rate fields, representing steady crustal deformation, using Global Navigation Satellite System (GNSS) velocity data. In the context of crustal deformation, PINNs have been applied to earthquake deformation analysis (Okazaki et al., 2022, 2024, 2025) and surface stress and strain fields modeling (Poulet and Behnoudfar, 2023, 2024). In this study, PDEs are used as physics-based regularization rather than as strict governing equations. Section 4 concludes the study with final remarks.

**Table 1.** Solution procedures of physics-informed machine learning (PIML) methods

| Method | Differentiation | Integration | Solution |
|---|---|---|---|
| PILM | Analytical | Analytical | Analytical |
| PIGP | Analytical | Collocation points | Analytical |
| PINN | Automatic | Collocation points | Stochastic optimization |

PI, physics-informed; LM, linear model; GP, Gaussian process; NN, neural network.

## 2. Formulation with Illustrative Examples

### *2.1 Damped oscillation*

Damped oscillation describes the time evolution $u(t)$ of a point mass. Because both the argument (time $t$) and solution (position $u$) are scalars, this system is suitable for presenting the basic concepts of the PILM. The equation of motion is given by the second-order ODE:

$$mu_{tt} + cu_t + ku = 0, \tag{3}$$

where $m$, $c$, and $k$ are the mass, damping coefficient, and elastic constant, respectively. The subscripts denote the derivatives: $u_t = \frac{du}{dt}$, $u_{tt} = \frac{d^2u}{dt^2}$. To uniquely determine the system, the initial position $u_0$ and velocity $v_0$ are specified as

$$u(0) = u_0,\ u_t(0) = v_0. \tag{4}$$

First, the unknown solution $u(t)$ is represented as a linear combination of $M$ fixed basis functions:

$$u(t;\mathbf{a}) = \mathbf{\Phi}(t)^{\mathrm{T}}\mathbf{a} = \sum a_i\phi_i(t). \tag{5}$$

Next, the loss function is expressed in terms of the model parameters $\mathbf{a} \in \mathbb{R}^M$. The initial condition (IC) loss is given by:

$$L_{\mathrm{IC}}(\mathbf{a}) = (u_0 - \mathbf{\Phi}(0)^{\mathrm{T}}\mathbf{a})^2 + (v_0 - \dot{\mathbf{\Phi}}(0)^{\mathrm{T}}\mathbf{a})^2 = (\mathbf{d} - \mathbf{Ha})^{\mathrm{T}}(\mathbf{d} - \mathbf{Ha}), \tag{6}$$

where $\mathbf{d} = (u_0, v_0)^{\mathrm{T}}$ and $\mathbf{H} = (\mathbf{\Phi}(0), \dot{\mathbf{\Phi}}(0))^{\mathrm{T}}$. The ODE loss is calculated as:

$$\begin{aligned} L_{\mathrm{ODE}}(\mathbf{a}) &= \int \left(m\ddot{\mathbf{\Phi}}^{\mathrm{T}}\mathbf{a} + c\dot{\mathbf{\Phi}}^{\mathrm{T}}\mathbf{a} + k\mathbf{\Phi}^{\mathrm{T}}\mathbf{a}\right)^{\mathrm{T}}\left(m\ddot{\mathbf{\Phi}}^{\mathrm{T}}\mathbf{a} + c\dot{\mathbf{\Phi}}^{\mathrm{T}}\mathbf{a} + k\mathbf{\Phi}^{\mathrm{T}}\mathbf{a}\right)dt \\ &= \int \mathbf{a}^{\mathrm{T}}\left[m^2\ddot{\mathbf{\Phi}}\ddot{\mathbf{\Phi}}^{\mathrm{T}} + c^2\dot{\mathbf{\Phi}}\dot{\mathbf{\Phi}}^{\mathrm{T}} + k^2\mathbf{\Phi}\mathbf{\Phi}^{\mathrm{T}} + mc\left(\ddot{\mathbf{\Phi}}\dot{\mathbf{\Phi}}^{\mathrm{T}} + \dot{\mathbf{\Phi}}\ddot{\mathbf{\Phi}}^{\mathrm{T}}\right) + mk\left(\ddot{\mathbf{\Phi}}\mathbf{\Phi}^{\mathrm{T}} + \mathbf{\Phi}\ddot{\mathbf{\Phi}}^{\mathrm{T}}\right) + ck\left(\dot{\mathbf{\Phi}}\mathbf{\Phi}^{\mathrm{T}} + \mathbf{\Phi}\dot{\mathbf{\Phi}}^{\mathrm{T}}\right)\right]\mathbf{a}dt \\ &= \mathbf{a}^{\mathrm{T}}[m^2\mathbf{R}^{22} + c^2\mathbf{R}^{11} + k^2\mathbf{R}^{00} + mc(\mathbf{R}^{21} + \mathbf{R}^{12}) + mk(\mathbf{R}^{20} + \mathbf{R}^{02}) + ck(\mathbf{R}^{10} + \mathbf{R}^{01})]\mathbf{a} \\ &= \mathbf{a}^{\mathrm{T}}\mathbf{Ga}. \end{aligned} \tag{7}$$

Here, we define the $M \times M$ definite-integral matrices of the basis functions as

$$\mathbf{R}^{ab} = \int \mathbf{\Phi}^{(a)}\mathbf{\Phi}^{(b)\mathrm{T}}dt, \tag{8}$$

where $a$ and $b$ denote the orders of the derivatives. Two points arise in the derivation: (i) since the ODE coefficients are constant, the integral applies only to $\mathbf{\Phi}(t)$; (ii) since the ODE is linear in $\mathbf{a}$, the ODE loss becomes a quadratic form. The total loss function is represented as

$$L(\mathbf{a}) = (\mathbf{d} - \mathbf{Ha})^{\mathrm{T}}(\mathbf{d} - \mathbf{Ha}) + \mathbf{a}^{\mathrm{T}}\mathbf{Ga}. \tag{9}$$

Therefore, the IC and ODE can be interpreted as the observation equation and regularization, respectively. This is a quadric function of $\mathbf{a}$, and its global minimum is given by

$$\mathbf{a}^* = \left(\mathbf{H}^{\mathrm{T}}\mathbf{H} + \mathbf{G}\right)^{-1}\mathbf{H}^{\mathrm{T}}\mathbf{d}. \tag{10}$$

If $\mathbf{H}$ and $\mathbf{R}$ are calculated analytically, then the optimal solution $u(t;\mathbf{a}^*)$ can be solved exactly. Notably, these are dependent not on the target PDEs and BCs but solely on the choice of basis functions; once these values are obtained, they can be used for other problems. In this study, we adopt cubic B-spline functions as basis functions, which are commonly used in geophysical inversion (e.g., Yabuki and Matsu'ura, 1992). We define a reference function that is non-zero in the interval $(-2, 2)$ as follows:

$$\bar{\phi}(x) = \frac{1}{6} \times \begin{cases} (x+2)^3 & -2 \le x \le -1 \\ (-x^3 - 6x^2 + 4) & -1 \le x \le 0 \\ (3x^3 - 6x^2 + 4) & 0 \le x \le 1 \\ -(x-2)^3 & 1 \le x \le 2 \end{cases}. \quad (11)$$

We then place $M$ basis functions over the interval $[0, T]$ as

$$\phi_i(t) = \bar{\phi}\left(\frac{t-(i-1)\Delta t}{\Delta t}\right) \left(\Delta t = \frac{T}{M-3}, i = 0, \dots, M-1\right). \quad (12)$$

Note that $\{\phi_0, \phi_1, \phi_2\}$ and $\{\phi_{M-1}, \phi_{M-2}, \phi_{M-3}\}$ are truncated at the domain boundaries $t = 0$ and $T$, respectively (Figure 1). Because they are piecewise cubic polynomials, the basis functions $\phi_i$ are twice continuously differentiable (Figure S1) and their definite integrals $\mathbf{R}^{ab}$ (eq. 8) can be calculated analytically. The values of $\mathbf{R}^{ab}$ are presented in Figure S2 and Table S1. Note that those for the case $a = b$ were provided by Nozue and Fukahata (2022).

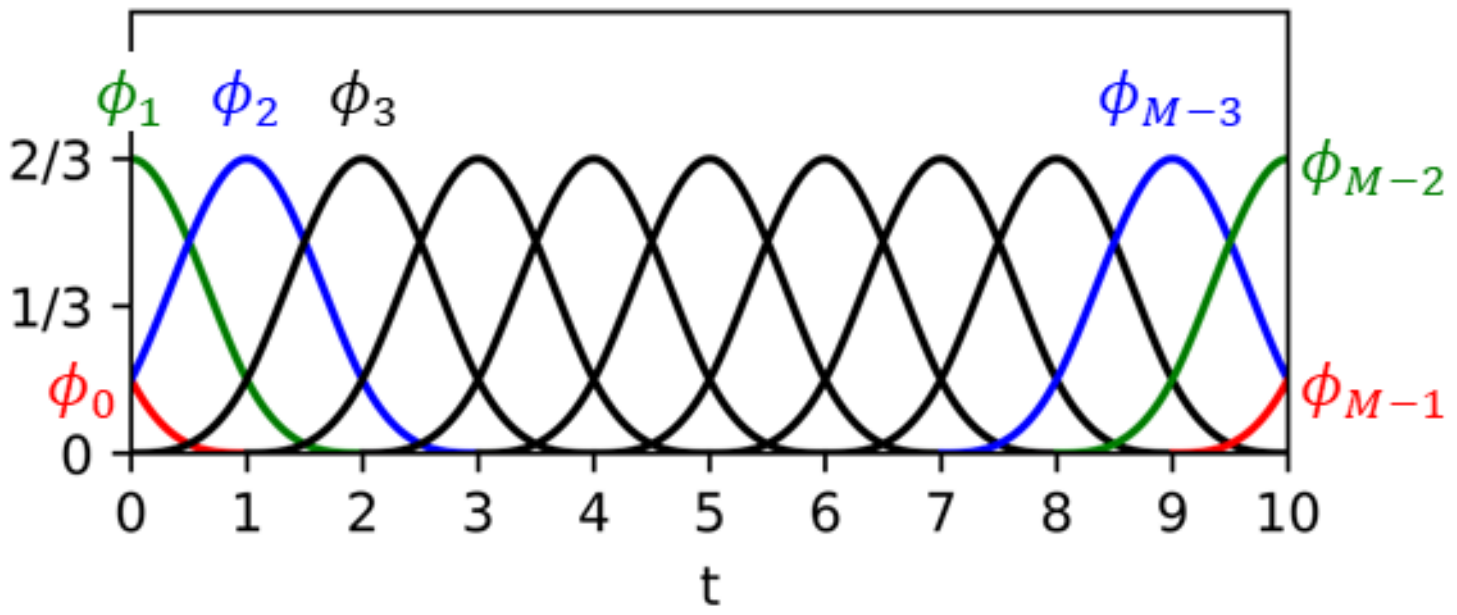

**Figure 1.** Shape of the basis functions for $T = 10$ and $M = 13$ ($\Delta t = 1$). Cubic B-spline functions are placed at equal intervals and are truncated at the domain boundaries.

First, the damped oscillation was solved for $T = 10$. The ODE coefficients were fixed at $m = 1$ and $k = 1$, and the damping coefficient was varied as $c = 0$ (harmonic oscillation), 1 (underdamping), 2 (critical damping), and 3 (overdamping). Three ICs, $(u_0, v_0) = (1,0)$, $(0.5, 0.5)$, and $(0, 1)$ were considered. The number of basis functions was set to $M = 13$ and 103, corresponding to intervals $\Delta t = 1$ and 0.1, respectively. The optimal solutions (eq. 10) are shown in Figure 2. The solutions are accurate, except for a visible discrepancy in the smaller model $M = 13$ (especially for $c = 0$ and 3), validating the PILM formulation.

Next, the scaling with respect to the number of model parameters ($M$) was empirically examined. The damped oscillation was solved for the coefficients $m = 1$, $c = 0$, and $k = 1$ (harmonic oscillation) with $T = 100$ under IC $(u_0, v_0) = (1,0)$. The optimal solutions for different values of $M$ are shown in Figure 3a. The results for $M = 103$ ($\Delta t = 1$) clearly exhibits a false decay in amplitude, although the phase is accurately captured. Notably, this is not an optimization failure as frequently observed in PINNs (e.g., Krishnapriyan et al., 2021; Wang et al., 2024) but rather a limitation in the model's representation capacity. As $M$ increases, the PILM solutions approach the analytical solution. The quantitative dependence was evaluated for $M = \{2^3, 2^4, \dots, 2^{11}\}$ as shown in Figure 3b. The memory consumption and computational time generally scale as $M^2$ and $M^3$, respectively. The loss functions and errors (i.e., difference in the PILM and analytical solutions) scales as $M^{-3}$ or better. The minimum eigen value of $\mathbf{G}$, which is zero if a model space contains a solution to ODEs, scales as $M^{-4}$. These results indicate promising scaling with respect to $M$. However, harmonic oscillation has a constant frequency and is relatively easy to solve with equally spaced basis functions. The performance may degrade for stiff ODEs with varying characteristic time scales, such as spring–slider systems (Fukushima et al., 2023).

Finally, inverse problems are addressed. Specifically, the damping coefficient $c$ (true value of 0.5) is estimated from discrete data $\mathbf{d} = (u_1, \dots, u_N)^\mathrm{T}$ at $(t_1, \dots, t_N)$. The true coefficients $m = 1$ and $k = 1$ are assumed known, but IC $(u_0, v_0) = (0.5, 0.5)$ is unknown. The loss function is the same as in eq. (9), with the observation matrix replaced by

$$\mathbf{H} = \begin{pmatrix} \phi_1(t_1) & \cdots & \phi_M(t_1) \\ \vdots & \ddots & \vdots \\ \phi_1(t_N) & \cdots & \phi_M(t_N) \end{pmatrix}. \tag{13}$$

The estimated parameters consist of both model parameters and an unknown coefficient: $(\mathbf{a},\ c)$. As matrix $\mathbf{G}$ depends on $c$, the minimum of the loss function $L(\mathbf{a}, c)$ cannot be solved analytically. Therefore, the optimal parameters $\mathbf{a}^*(c)$ are obtained from eq. (10) as a function of $c$, and the optimal coefficient $c^*$ is determined by minimizing $L^*(c) = L(\mathbf{a}^*(c), c)$ using a grid search. In numerical experiments, ten data points were generated at $t = (0.5, 1.5, \dots, 9.5)$ by adding independent Gaussian noise to the analytical solutions. Four datasets were prepared with noise standard deviation $\sigma$ of 0, 0.01, 0.05, and 0.1. The results are presented in Figure 4. The estimated damping coefficients $c^*$ are 0.50, 0.51, 0.51, and 0.34, respectively (Figure 4a). For noise-free data, the loss function exhibits a sharp minimum at the true value. The estimates remain accurate at low noise levels, although the minimum loss value is larger. For the highest noise level, the estimate becomes inaccurate; the underestimation may result from an upward noise at a local maximum at $t = 7.5$ (Figure 4b). The PILM provides plausible estimations for inverse problems.

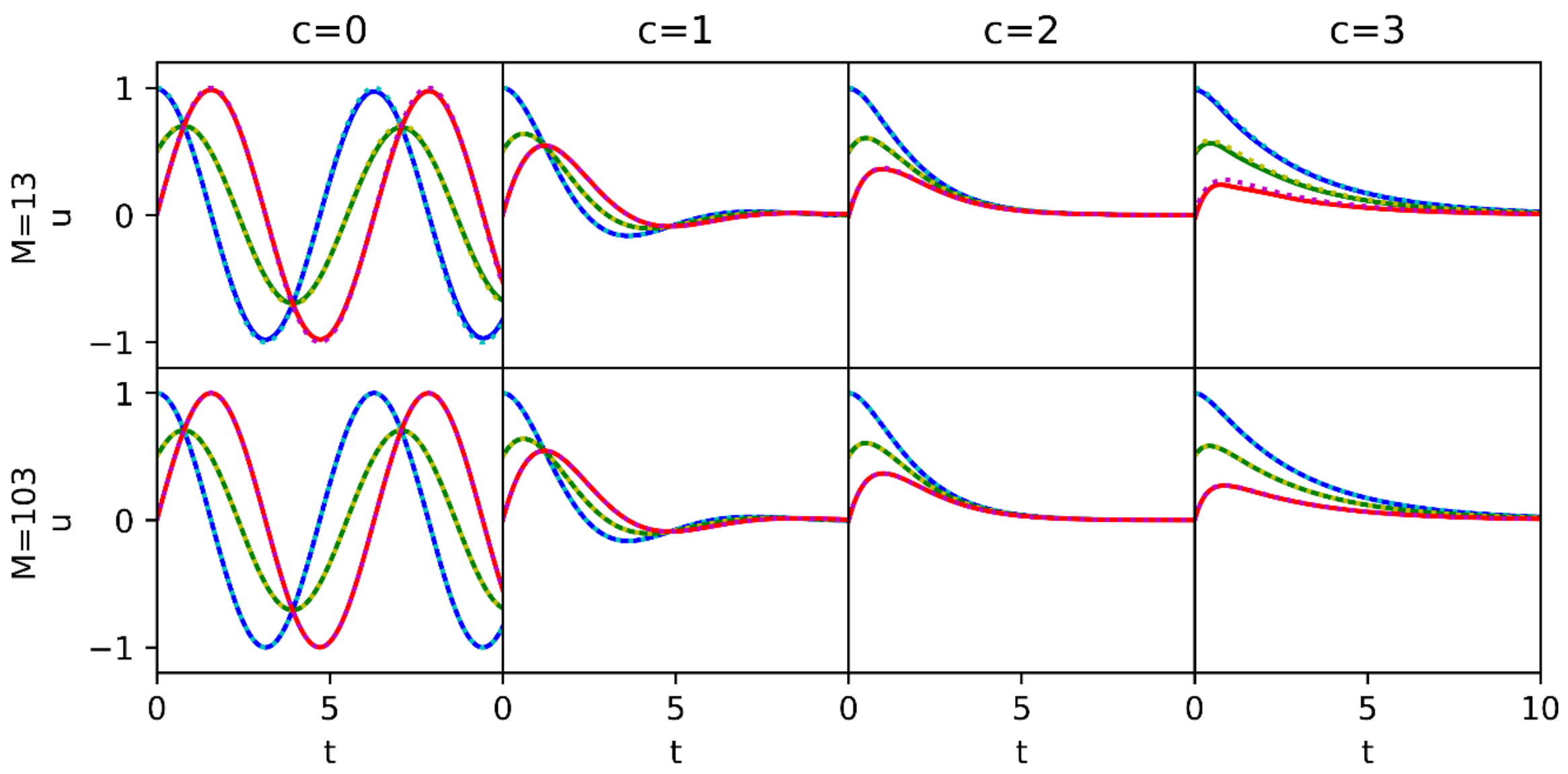

**Figure 2.** PILM solutions to damped oscillation with $m = 1$ and $k = 1$ for different damping coefficients ($c$) and number of basis functions ($M$). Blue, green, and red lines represent initial positions and velocities of $(u_0, v_0) = (1,0)$, $(0.5,\ 0.5)$, and $(0,\ 1)$, respectively. Solid and dashed lines indicate PILM and analytical solutions, respectively.

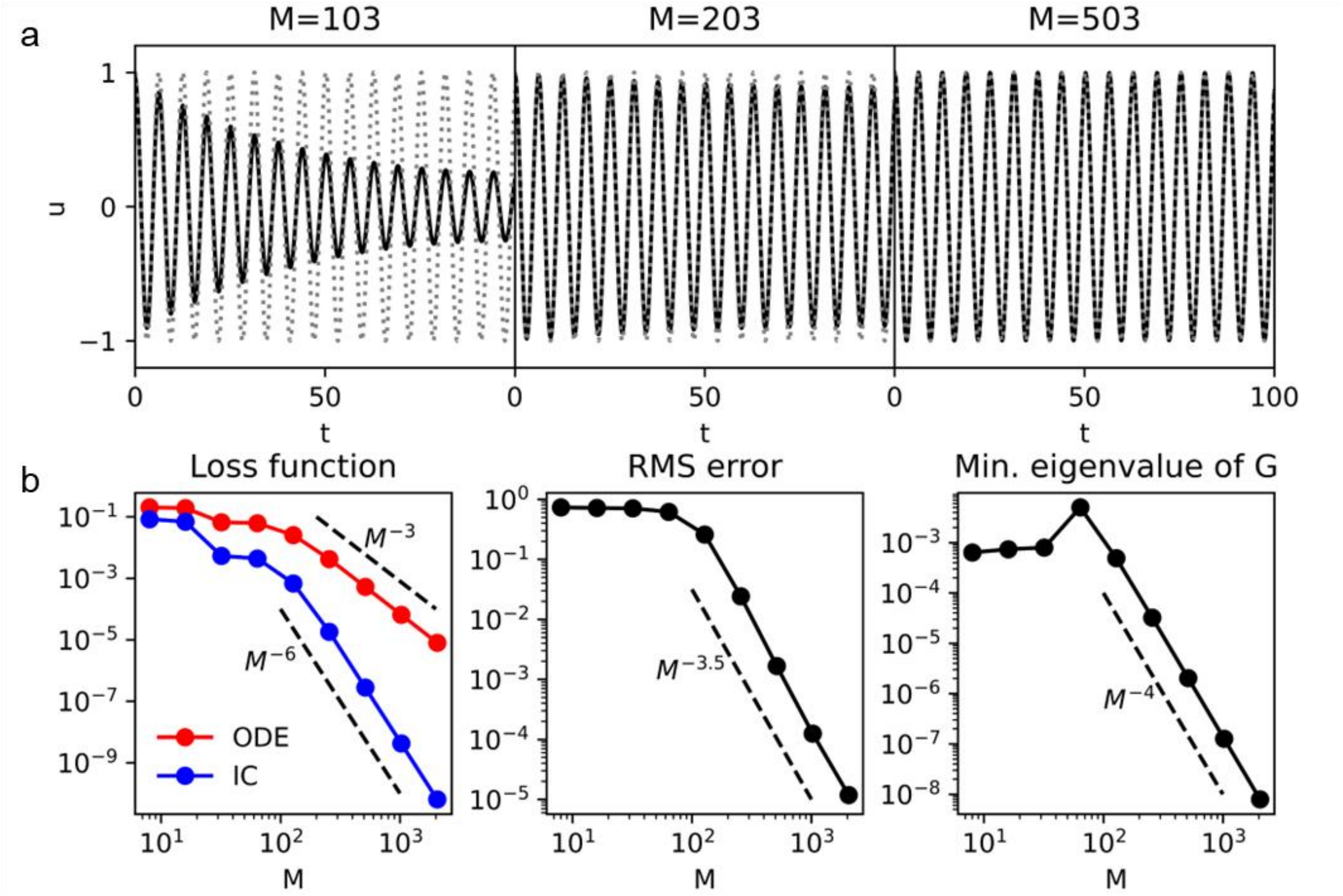

**Figure 3.** Dependence of PILM solutions on the number of basis functions ($M$). (a) PILM solutions to damped oscillation with $m = 1$, $c = 0$, and $k = 1$ for different values of $M$. Solid and dashed lines indicate PILM and analytical solutions, respectively. (b) Scaling law of the loss functions, root-mean-square (RMS) error of the solutions, and the minimum eigenvalue of the matrix $\mathbf{G}$.

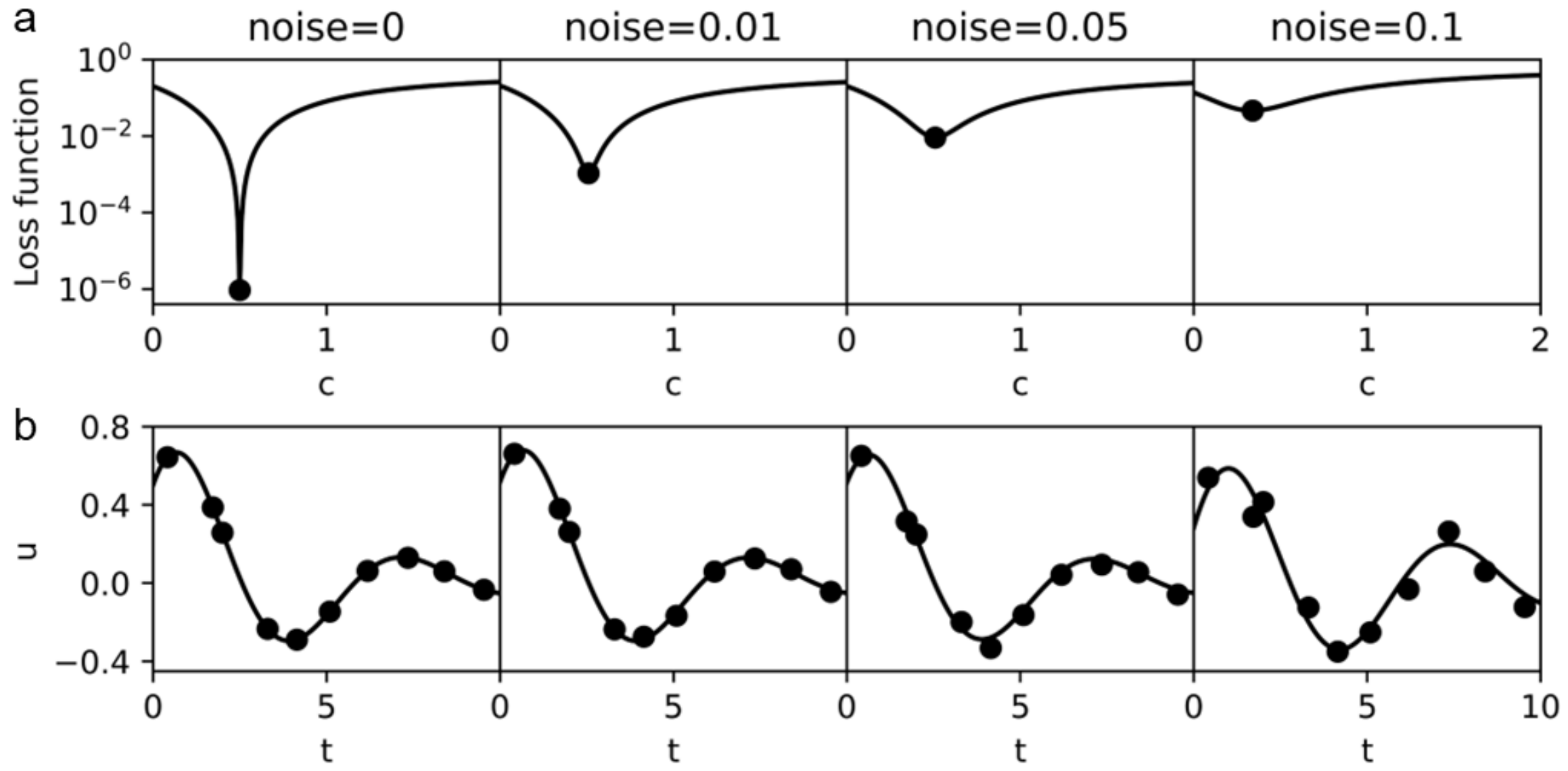

**Figure 4.** Inverse analysis of the damped oscillation using PILM. (a) Loss function $L^*(c)$ as a function of assumed damping coefficient $c$. Black dots indicate the optimal coefficients $c^*$. (b) Optimal solutions for the estimated $c^*$. Lines and dots represent the PILM estimation and observed data, respectively.

### *2.2 Diffusion equation*

We consider the diffusion equation in one spatial dimension. The temperature field $u(t,x)$ obeys the following second-order PDE:

$$\frac{du}{dt} - k\frac{d^2u}{dx^2} = 0, \quad (14)$$

where $k$ is the diffusion coefficient. This serves as a simple PDE system with multiple input variables. We define the basis functions $\phi_i(t,x)$ as the product of cubic B-spline functions in the $t$ and $x$ directions and arrange in a lattice pattern:

$$\phi_i(t,x) = T_i(t)X_i(x) = \bar{\phi}\left(\frac{t-(k-1)\Delta t}{\Delta t}\right)\bar{\phi}\left(\frac{x-(l-1)\Delta x}{\Delta x}\right)\ \left(\Delta t = \frac{T}{M_t-3}, \Delta x = \frac{X}{M_x-3}\right), \quad (15)$$

for $k = 0, \ldots, M_t - 1$, $l = 0, \ldots, M_x - 1$, and $i = kM_x + l = 0, \ldots, M - 1$ with $M = M_t M_x$. The PDE loss is expressed as

$$\begin{aligned} L_{\mathrm{PDE}}(\mathbf{a}) &= \int \sum a_i a_j (T_i' X_i - kT_i X_i'')(T_j' X_j - kT_j X_j'')dtdx \\ &= \sum a_i a_j \Big[\int T_i' T_j' dt \int X_i X_j dx + k^2 \int T_i T_j dt \int X_i'' X_j'' dx - k\Big(\int T_i' T_j dt \int X_i X_j'' dx + \\ &\int T_i T_j' dt \int X_i'' X_j dx\Big)\Big] \\ &= \mathbf{a}^{\mathrm{T}}[\mathbf{R}^{11} \otimes \mathbf{R}^{00} + k^2 \mathbf{R}^{00} \otimes \mathbf{R}^{22} - k(\mathbf{R}^{10} \otimes \mathbf{R}^{02} + \mathbf{R}^{01} \otimes \mathbf{R}^{20})]\mathbf{a} \\ &= \mathbf{a}^{\mathrm{T}}\mathbf{G}\mathbf{a}. \end{aligned} \quad (16)$$

Here, $\otimes$ represents the tensor (Kronecker) product of matrices, where the left and right sides correspond to $T_i(t)$ and $X_i(x)$, respectively. Therefore, matrices $\mathbf{R}^{ab}$ in one dimension (eq. 8) can be used to calculate $\mathbf{G}$ in two dimensions. In general, $\mathbf{G}$ of arbitrary second-order linear PDEs can be constructed from $\mathbf{R}^{ab}$. This universality in PDE forms and dimensions is a practical advantage of the PILM formulation.

As a numerical experiment, the following inverse problem was considered. The temperature field obeys the diffusion equation with $k = 0.1$ and is measured at equally spaced sensors at constant time intervals (Figure 5). However, the diffusion coefficient $k$, BC, and IC are unknown. We attempted to estimate $k$ and reconstruct the initial temperature $u(0,x)$ from measurements at $t > 0$. To solve this problem, we estimated $k^*$ by numerically minimizing the optimal loss $L^*(k)$, which is obtained from the least-squares solutions for a given $k$. Four datasets were prepared for inversions, corresponding to unimodal and bimodal initial distributions with clean or noisy measurements. The number of measurements differs between unimodal and bimodal distributions, considering the complexity of the fields.

The results are presented in Figure 6. The estimated diffusion coefficients $k^*$ are 0.100, 0.100, 0.095, and 0.096 for the unimodal (clean), bimodal (clean), unimodal (noisy), and bimodal (noisy) datasets, respectively (Figure 6a). These are quite accurate and robust to noise. This is because $k$ can be estimated from the entire domain, including the later diffusion times. The data fitting and initial value reconstruction are shown in Figure 6b. The estimated solutions fit the clean data well, whereas some differences are observed for noisy data. The PDE loss effectively functions as regularization to suppress overfitting. The initial reconstruction succeeded in the unimodal distribution with clean measurements but failed in the other cases. Owing to the smoothing effect of the diffusion equation (i.e., rapid information loss over time), small errors lead to significant discrepancies in the initial distributions. Although BCs and ICs must typically be specified or assumed in numerical solvers, they were not required in this experiment. This illustrates the potential of PILM for inverse problems involving uncertain BCs and ICs.

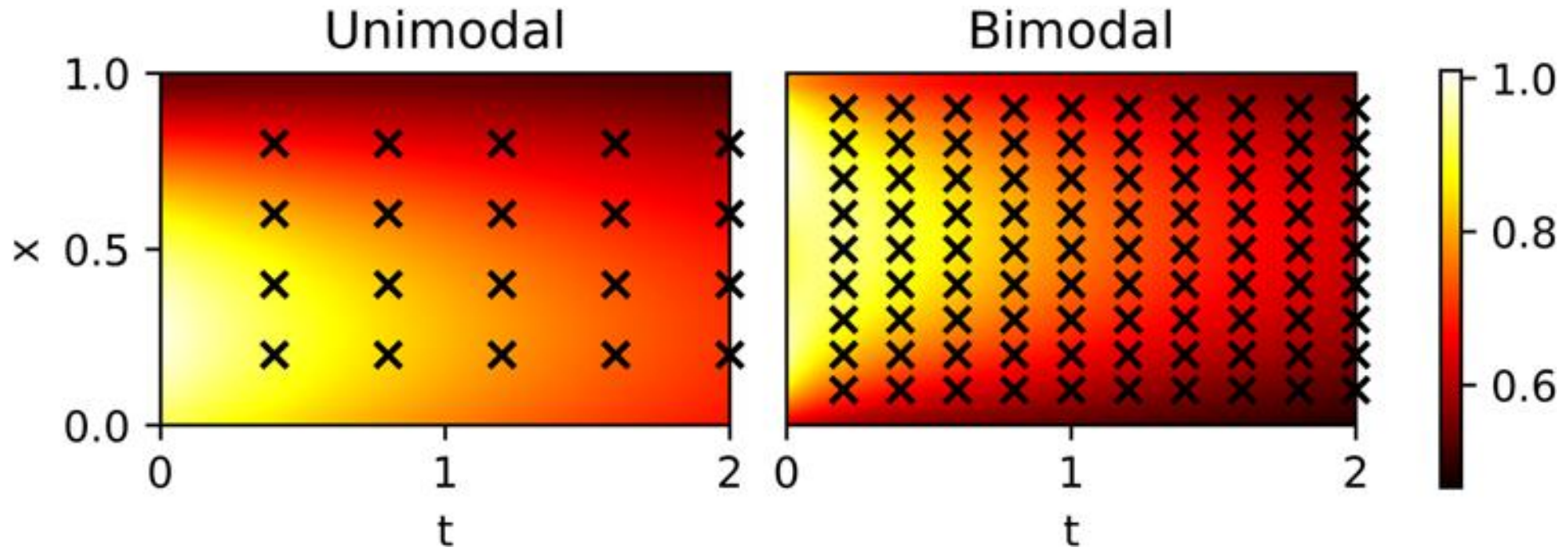

**Figure 5.** Heatmap of true solutions to the diffusion equation. Cross marks indicate measurement points.

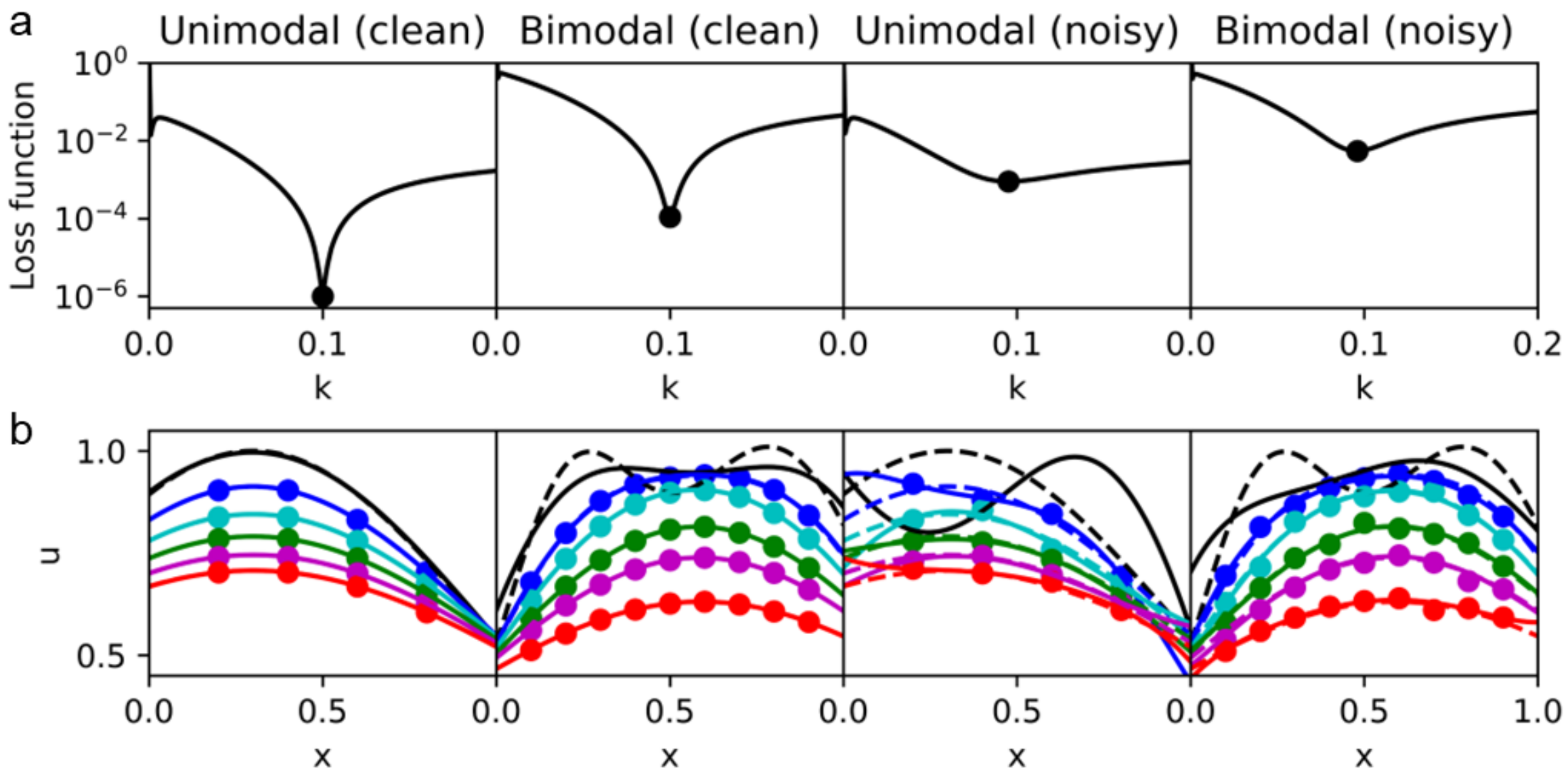

**Figure 6.** Inverse analysis of the diffusion equation using PILM. (a) Loss function $L^*(k)$ as a function of the assumed diffusion coefficient $k$. Black dots indicate the optimal coefficients $k^*$. (b) Optimal solutions for the estimated $k^*$. Solid and dashed lines represent PILM estimation and true solution, respectively. Dots represent data. Black lines indicate the initial temperature. Blue, cyan, green, magenta, and red lines correspond to $t = 0.4$, 0.8, 1.2, 1.6, and 2.0 for the unimodal solution and $t = 0.2$, 0.4, 0.8, 1.2, and 2.0 for the bimodal solution.

## 3. Application to Crustal Strain Rate Estimation

Quantitative evaluation of steady crustal deformation provides fundamental insights into tectonic processes such as earthquake preparation processes. Geodetic observations have enabled precise estimation of regional strain rate fields (i.e., the spatial derivatives of velocity fields). Since the deployment of GNSS observation networks around the 1990s, methods for inverting continuous strain rate fields from discrete GNSS velocity data have been actively investigated (e.g., Haines and Holt, 1993; Shen et al., 1996; Tape et al., 2009) and continue to be developed (e.g., Pagani et al., 2021; Maurer and Materna, 2023).

The basis function expansion method was applied to leveling data by Fukahata et al. (1996) and GNSS data by Okazaki et al. (2021), and has been used to estimate regional strain-rate fields (e.g., Nishimura, 2022; Nozue and Fukahata, 2025). These studies imposed smoothness on the velocity fields to ensure stable estimations. In contrast, Sandwell and Wessel (2016) proposed a velocity interpolation method using elasticity constraints, which may more realistically describe the force balance in the crust. In this section, using the PILM framework, physical regulation (i.e., elasticity constraints) is incorporated into the basis function expansion method and compared with mathematical regularization (i.e., smoothing constraints).

### *3.1 Mathematical and physical regularization*

This task is a typical regression problem for estimating a continuous horizontal velocity field $(u, v)$ as a function of the position $(x, y)$ from discrete observational data $(x_i, y_i, u_i, v_i)_{i=1}^{N}$. The strain rate is obtained from spatial differentiation of the estimated velocity field. Each velocity component is expanded using the basis functions as:

$$u = \mathbf{\Phi}(x,y)^{\mathrm{T}}\mathbf{a}_x = \sum a_{xi} X_i(x) Y_i(y),\ \ v = \mathbf{\Phi}(x,y)^{\mathrm{T}}\mathbf{a}_y = \sum a_{yi} X_i(x) Y_i(y),\ \ \mathbf{a} = \begin{pmatrix}\mathbf{a}_x \\ \mathbf{a}_y\end{pmatrix}. \tag{17}$$

Observation matrix $\mathbf{H}$ is the same as that in eq. (13), with a straightforward extension to two components.

In regression analysis, regularization based on prior knowledge is commonly introduced to suppress overfitting. Mathematical regularization, such as damping, smoothing, and sparsity constraints, is typically used in geodetic inversions (Nakata et al., 2016). Here, we assume smoothness by penalizing the second-order derivatives of the velocity field $(u, v)$ (i.e., spatial variations in the strain rate field), following previous studies (Yabuki and Matsu'ura, 1992; Okazaki et al., 2021). The mathematical regularization is calculated as

$$\begin{aligned} L_{\mathrm{math}}(\mathbf{a}) &= \int \left(u_{xx}^2 + u_{xy}^2 + u_{yx}^2 + u_{yy}^2 + v_{xx}^2 + v_{xy}^2 + v_{yx}^2 + v_{yy}^2\right) dxdy \\ &= \left(\mathbf{a}_x^{\mathrm{T}}\ \mathbf{a}_y^{\mathrm{T}}\right)\begin{pmatrix}\mathbf{G}' & \mathbf{O} \\ \mathbf{O} & \mathbf{G}'\end{pmatrix}\begin{pmatrix}\mathbf{a}_x \\ \mathbf{a}_y\end{pmatrix} = \mathbf{a}^{\mathrm{T}}\mathbf{G}_{\mathrm{math}}\mathbf{a}, \end{aligned} \tag{18}$$

with

$$\mathbf{G}' = \mathbf{R}^{22} \otimes \mathbf{R}^{00} + 2\mathbf{R}^{11} \otimes \mathbf{R}^{11} + \mathbf{R}^{00} \otimes \mathbf{R}^{22}. \tag{19}$$

Because $\mathbf{G}_{\mathrm{math}}$ is block-diagonal, the two velocity components are interpolated independently.

Regularization can also be defined based on physical requirements or consistency. Here, we assume that the crust deforms as a thin elastic sheet in a mechanically natural way (i.e., under the smallest possible external force). The equilibrium of inplane deformation under an external force $\mathbf{f}$ is given by a system of second-order PDEs (Sandwell and Wessel, 2016):

$$\begin{pmatrix}A u_{xx} + B v_{xy} + u_{yy} \\ A v_{yy} + B u_{xy} + v_{xx}\end{pmatrix} = -\frac{\mathbf{f}}{\mu}. \tag{20}$$

Heere, $A = \frac{2}{1-\nu}$ and $B = \frac{1+\nu}{1-\nu}$, where $\mu$ and $\nu$ are shear modulus and Poisson's ratio, respectively. We impose a penalty on the strength of the external force $\mathbf{f}$. Based on the PILM formulation, the physical regularization is calculated as

$$\begin{aligned} L_{\mathrm{phys}}(\mathbf{a}) &= \int \left(\frac{\mathbf{f}}{\mu}\right)^2 dxdy = \int \left[\left(A u_{xx} + B v_{xy} + u_{yy}\right)^2 + \left(A v_{yy} + B u_{xy} + v_{xx}\right)^2\right] dxdy \\ &= \left(\mathbf{a}_x^{\mathrm{T}}\ \mathbf{a}_y^{\mathrm{T}}\right)\begin{pmatrix}\mathbf{G}_{xx} & \mathbf{G}_{xy} \\ \mathbf{G}_{yx} & \mathbf{G}_{yy}\end{pmatrix}\begin{pmatrix}\mathbf{a}_x \\ \mathbf{a}_y\end{pmatrix} = \mathbf{a}^{\mathrm{T}}\mathbf{G}_{\mathrm{phys}}\mathbf{a}, \end{aligned} \tag{21}$$

with

$$\mathbf{G}_{xx} = A^2\mathbf{R}^{22}\otimes\mathbf{R}^{00} + A(\mathbf{R}^{20}\otimes\mathbf{R}^{02} + \mathbf{R}^{02}\otimes\mathbf{R}^{20}) + \mathbf{R}^{00}\otimes\mathbf{R}^{22} + B^2\mathbf{R}^{11}\otimes\mathbf{R}^{11},$$

$$\mathbf{G}_{xy} = B[A(\mathbf{R}^{21}\otimes\mathbf{R}^{01} + \mathbf{R}^{10}\otimes\mathbf{R}^{12}) + \mathbf{R}^{01}\otimes\mathbf{R}^{21} + \mathbf{R}^{12}\otimes\mathbf{R}^{10}],$$

$$\mathbf{G}_{yx} = B[A(\mathbf{R}^{12}\otimes\mathbf{R}^{10} + \mathbf{R}^{01}\otimes\mathbf{R}^{21}) + \mathbf{R}^{10}\otimes\mathbf{R}^{12} + \mathbf{R}^{21}\otimes\mathbf{R}^{01}],$$

$$\mathbf{G}_{yy} = A^2\mathbf{R}^{00}\otimes\mathbf{R}^{22} + A(\mathbf{R}^{02}\otimes\mathbf{R}^{20} + \mathbf{R}^{20}\otimes\mathbf{R}^{02}) + \mathbf{R}^{22}\otimes\mathbf{R}^{00} + B^2\mathbf{R}^{11}\otimes\mathbf{R}^{11}. \quad (22)$$

Physical regularization induces a free parameter $\nu$ taking values in $[-1, 0.5]$. Due to the off-block-diagonal components $\mathbf{G}_{xy}$ and $\mathbf{G}_{yx}$, the two velocity components are generally coupled. If $\nu = -1$ (i.e., $B = 0$) is specified, $\mathbf{G}_{\text{phys}}$ is block-diagonal and the two components are decoupled. The derived expressions are numerically verified in Figures S3 and S4.

### *3.2 Bayesian formulation*

We introduce the Bayesian framework adopted in previous studies (Yabuki and Matsu'ura, 1992; Okazaki et al., 2021). Regularization is regarded as a prior distribution of the model parameters:

$$p(\mathbf{a}) \propto \exp\left[-\frac{1}{2\rho_i^2}\mathbf{a}^{\mathrm{T}}\mathbf{G}_i\mathbf{a}\right], \quad (23)$$

where subscripts $i$ denote 'math' or 'phys', $\rho_i$ is a hyperparameter describing the strength of prior knowledge, and the symbol $\propto$ indicates that the normalization constant is omitted. By defining the likelihood as

$$p(\mathbf{d}|\mathbf{a}) \propto \exp\left[-\frac{1}{2\sigma^2}(\mathbf{d} - \mathbf{Ha})^{\mathrm{T}}(\mathbf{d} - \mathbf{Ha})\right], \quad (24)$$

where $\sigma$ is a hyperparameter representing standard deviation of observational noise, a posterior distribution is given by

$$p(\mathbf{a}|\mathbf{d}) \propto p(\mathbf{d}|\mathbf{a})p(\mathbf{a}) \propto \exp\left[-\frac{1}{2\sigma^2}\left\{(\mathbf{d} - \mathbf{Ha})^{\mathrm{T}}(\mathbf{d} - \mathbf{Ha}) + \alpha_i^2\mathbf{a}^{\mathrm{T}}\mathbf{G}_i\mathbf{a}\right\}\right], \quad (25)$$

where $\alpha_i^2 = \sigma^2/\rho_i^2$ is the hyperparameter describing the relative importance of data and prior knowledge. The maximum a posteriori (MAP) estimation $\mathbf{a}_*$ that maximizes $p(\mathbf{a}|\mathbf{d})$ is obtained as

$$\mathbf{a}_i^* = \left(\mathbf{H}^{\mathrm{T}}\mathbf{H} + \alpha_i^2\mathbf{G}_i\right)^{-1}\mathbf{H}^{\mathrm{T}}\mathbf{d}. \quad (26)$$

This is equivalent to the least-squares solution of the PILM, whose loss function is given by

$$L_i(\mathbf{a}) = (\mathbf{d} - \mathbf{Ha})^{\mathrm{T}}(\mathbf{d} - \mathbf{Ha}) + \alpha_i^2\mathbf{a}^{\mathrm{T}}\mathbf{G}_i\mathbf{a}, \quad (27)$$

and $\alpha_i$ corresponds to the relative weight between the loss terms.

The benefit of a Bayesian formulation lies in the objective selection of hyperparameter values, that is, the adaptive weights of the loss terms in PIML (e.g., Wang et al., 2021, 2022). We adopt the maximum marginal likelihood criterion. The log marginal likelihood can be analytically maximized with respect to $\sigma^2$ and be expressed as a function of $\alpha_i^2$ (Yabuki and Matsu'ura, 1992):

$$\sigma^{*2} = \frac{L_i(\mathbf{a}_i^*)}{N - M + P_i}, \quad (28)$$

$$\mathrm{LL}\left(\alpha_i^2; \mathbf{G}_i\right) = \frac{1}{2}\left[-(N - M + P_i)(\log(2\pi\sigma^{*2}) + 1) + P_i\log\alpha_i^2 + \log\|\mathbf{G}_i\| - \log\left\|\mathbf{H}^{\mathrm{T}}\mathbf{H} + \alpha_i^2\mathbf{G}_i\right\|\right], \quad (29)$$

where $P_i$ and $\|\mathbf{G}_i\|$ are the rank and the product of the non-zero eigenvalues of $\mathbf{G}_i$, respectively. An optimal pair $(\alpha_i^{*2}, \text{LL}_i^*)$ is explored using a grid search.

### *3.3 Data and results*

Okazaki et al. (2021) processed the GNSS horizontal velocities in and around Japan during 2006–2009, which were archived by the Geospatial Information Authority of Japan, the Japan Coast Guard, Kyoto University, the International GNSS service, and UNAVCO. In this study, a subset of this dataset was used, which is a part of central Japan defined as a 400 × 400 km square centered at (138°E, 36°N), leading to 458 GNSS stations (Figure 7). Cubic B-spline functions were placed at 20 km intervals, yielding 529 parameters for each velocity component. In physical regularization, Poisson's ratio of 0.5, 0, and −1 was considered.

Graphs of the log marginal likelihood are shown in Figure 8a, and the statistics of the optimal hyperparameters are presented in Table 2. The log marginal likelihood indicates the clear superiority of mathematical regularization. The optimal noise standard deviation ($\sigma^*$) and root-mean-square error on the velocity data are smaller for mathematical regularization, indicating higher fitting accuracy. The estimated maximum shear strain rate fields are shown in Figure 9. The result of mathematical regularization resolves high shear zones around Niigata ($x \sim 0, y \sim 100$ km) and Izu Peninsula ($x \sim 100$, $y \sim -100$ km), which were observed in previous studies (e.g., Kato et al., 1998; Sagiya et al., 2000). However, these features are blurred in the physical regularization. They are slightly visible in the decoupled model ($\nu = -1$) but are barely visible in the coupled models ($\nu = 0.5$ and 0). Furthermore, extreme strain rates are estimated at the margins of the model region. Although the estimates in the ocean (areas without stations) are not considered real, artificial high strain rates also appear on land at the top-right corner. In summary, the mathematical regularization exhibits higher resolution and stability, suggesting that elastic coupling can degrade the estimation of crustal strain rates.

In this study, the external forces in the physical regularization were continuously distributed across the model domain. In contrast, Sandwell and Wessel (2016) assumed that point forces acted at the locations of velocity measurements, which could resolve deformation in densely observed areas and suppress artificial deformation in the ocean. Although the physical correspondence of such localized forces is unclear, the method of Sandwell and Wessel (2016) would produce a more reasonable estimation than the physical regularization presented here.

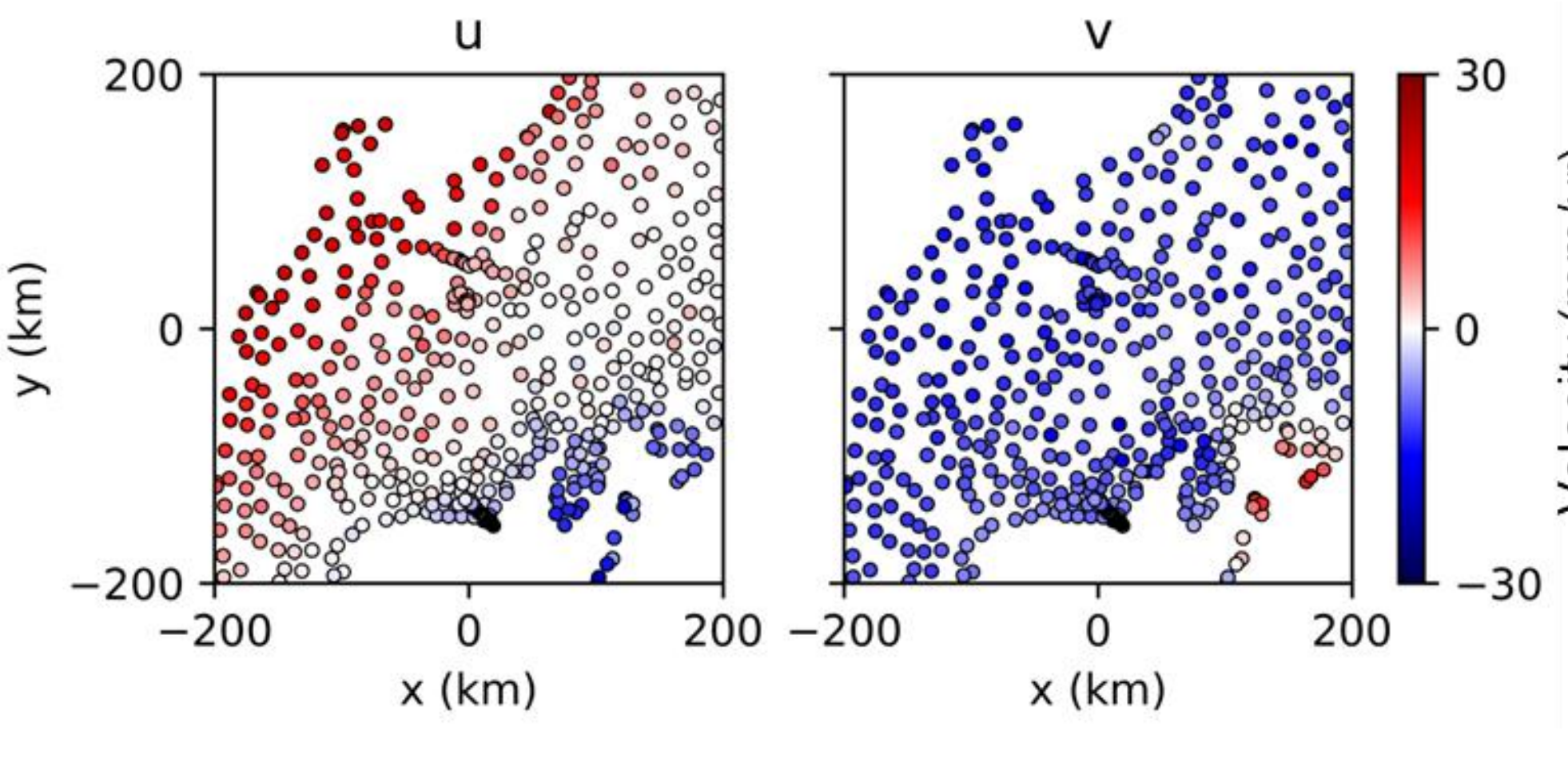

**Figure 7.** Velocity data from 458 GNSS stations in central Japan. The model region is defined as a 400 × 400 km square centered at (138°E, 36°N).

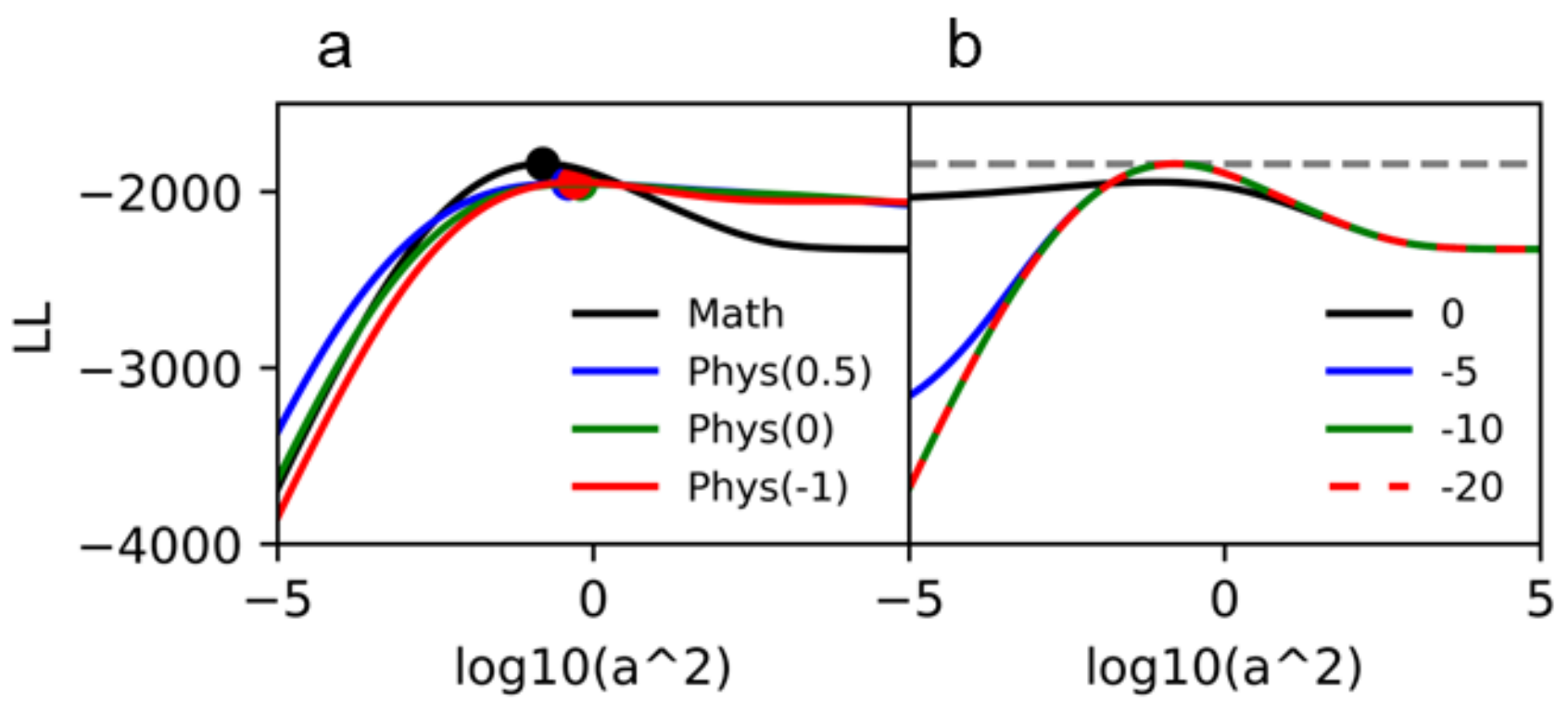

**Figure 8.** Log marginal likelihood plots for (a) mathematical and physical regularization ($\nu = 0.5$, 0, and $-1$) and (b) hybrid regularization as a function of $\log_{10} \alpha^2_{\text{math}}$ with fixed values of $\log_{10} \alpha^2_{\text{phys}} = 0, -5, -10$, and $-20$. The dashed gray line indicates the optimal value under mathematical regularization.

**Table 2.** Optimal hyperparameter values and fitting accuracy for different regularization methods

| Regularization | Log marginal likelihood | Noise standard deviation (mm/yr) | Root-mean-square error (mm/yr) |
|---|---|---|---|
| Mathematical | −1842.75 | 1.42 | 1.26 |
| Physical ($\nu = 0.5$) | −1954.48 | 1.67 | 1.55 |
| Physical ($\nu = 0$) | −1956.03 | 1.62 | 1.50 |
| Physical ($\nu = -1$) | −1947.09 | 1.53 | 1.38 |

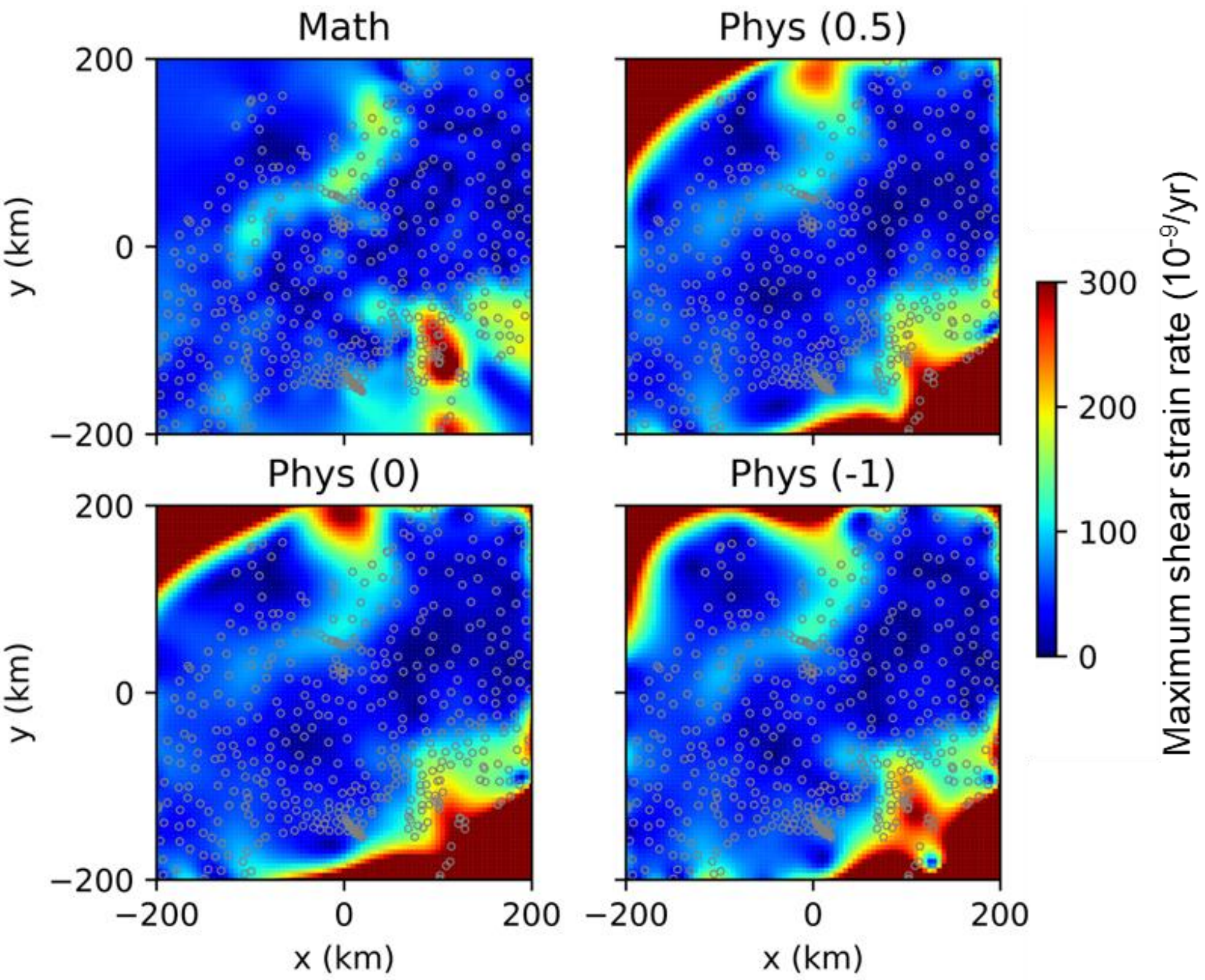

**Figure 9.** Maximum shear strain rate fields estimated using mathematical and physical regularization ($\nu = 0.5$, 0, and $-1$). Grey circles indicate GNSS stations.

### *3.4 Hybrid regularization*

Although physical regularization yields a lower marginal likelihood than mathematical regularization, an appropriate combination of the two may achieve higher marginal likelihood. Therefore, we examine a hybrid of the two regularization terms. We define the prior distribution of $\mathbf{a}$ by (Fukahata et al., 2004)

$$p(\mathbf{a}) \propto \exp\left[-\frac{1}{2\rho_{\mathrm{math}}^2}\mathbf{a}^{\mathrm{T}}\mathbf{G}_{\mathrm{math}}\mathbf{a} - \frac{1}{2\rho_{\mathrm{phys}}^2}\mathbf{a}^{\mathrm{T}}\mathbf{G}_{\mathrm{phys}}\mathbf{a}\right]. \quad (30)$$

The MAP solution is given by

$$\mathbf{a}_{\mathrm{hyb}}^* = \left(\mathbf{H}^{\mathrm{T}}\mathbf{H} + \alpha_{\mathrm{math}}^2\mathbf{G}_{\mathrm{math}} + \alpha_{\mathrm{phys}}^2\mathbf{G}_{\mathrm{phys}}\right)^{-1}\mathbf{H}^{\mathrm{T}}\mathbf{d}. \quad (31)$$

Considering the identity

$$\frac{1}{\rho_{\mathrm{math}}^2}\mathbf{G}_{\mathrm{math}} + \frac{1}{\rho_{\mathrm{phys}}^2}\mathbf{G}_{\mathrm{phys}} = \frac{1}{\rho_{\mathrm{math}}^2}\left(\mathbf{G}_{\mathrm{math}} + \frac{\alpha_{\mathrm{phys}}^2}{\alpha_{\mathrm{math}}^2}\mathbf{G}_{\mathrm{phys}}\right) = \frac{1}{\rho_{\mathrm{phys}}^2}\left(\mathbf{G}_{\mathrm{phys}} + \frac{\alpha_{\mathrm{math}}^2}{\alpha_{\mathrm{phys}}^2}\mathbf{G}_{\mathrm{math}}\right), \quad (32)$$

we define matrices

$$\widetilde{\mathbf{G}}_i = \mathbf{G}_i + \frac{\alpha_{-i}^2}{\alpha_i^2}\mathbf{G}_{-i}, \quad (33)$$

where $-i$ denotes 'math' or 'phys' such that $\{i, -i\} = \{\mathrm{math}, \mathrm{phys}\}$ holds as a set. Note that $\widetilde{\mathbf{G}}_i$ can be regarded as perturbations of $\mathbf{G}_i$ if $\alpha_i^2 \gg \alpha_{-i}^2$. Using this notation, the log marginal likelihood of hybrid regularization is represented as

$$\mathrm{LL}(\alpha_{\mathrm{math}}^2, \alpha_{\mathrm{phys}}^2; \mathbf{G}_{\mathrm{math}}, \mathbf{G}_{\mathrm{phys}}) = \mathrm{LL}\left(\alpha_i^2; \widetilde{\mathbf{G}}_i\right). \quad (34)$$

Although hybrid regularization was applied to the GNSS data in Section 3.3, the log marginal likelihood (34) for any combination of hyperparameter values $(\alpha_{\mathrm{math}}^2, \alpha_{\mathrm{phys}}^2)$ did not exceed $\mathrm{LL}_{\mathrm{math}}^*$ (Figure 8b). This result suggests that physical regularization is unsuitable for this dataset from a statistical perspective.

This study adopted the marginal likelihood for hyperparameter optimization, demonstrating the marginal superiority of mathematical over hybrid regularization, despite difference in the numbers of hyperparameters. This suggests that marginalization can effectively account for differences in model complexity (Bishop, 2006). In previous studies (Yabuki and Matsu'ura, 1992; Okazaki et al., 2021), Akaike's Bayesian information criterion (Akaike, 1980), defined by $\mathrm{ABIC} = -2\mathrm{LL} + 2N_h$ ($N_h$ is the number of hyperparameters), was minimized for hyperparameter optimization. Under this criterion, due to the penalty associated with an additional hyperparameter, mathematical regularization performs better than hybrid regularization by a finite margin.

## 4. Concluding Remarks

This study presented the analytical formulation and applications of PILM. In addition to typical forward and inverse problems (Section 2.1), estimation in underdetermined conditions (Section 2.2) and Bayesian linear regression with physical regularization (Section 3) were addressed. These diverse problems can be solved using a unified method that incorporates available physical information (PDEs, BCs, and ICs) and data into the loss functions. This flexibility is a salient feature of PIML approaches, including PILM.

A key advantage of PILM among the PIML approaches is its closed-form solution. Failure in PINN modeling can result from the limited representation ability of neural networks, approximation of loss functions (i.e., sampling collocation points), and difficulties in optimization. Isolating these potential causes is occasionally problematic. In contrast, the PILM does not suffer from approximation and optimization issues, allowing a focus on basis function design to improve accuracy. However, the PILM formulation in this study is limited to linear PDEs with fixed coefficients within rectangular input domains. Furthermore, the basis function expansion method is feasible only in low-dimensional problems due to computational costs. Therefore, if a target problem involves nonlinear equations, varying coefficients, complex geometries, or high dimensions, a versatile PINN approach would be a practical choice.

Finally, we note a close similarity to the finite element method (FEM), which also represents unknown solutions to PDEs as linear combinations of fixed basis functions and seeks optimal coefficients by minimizing an objective function defined as an integral over the PDE domain. A crucial difference is that the objective function in PILM consists of residuals from PDEs and BCs in strong form, whereas that in FEM is based on variational or weak form of the PDEs. Connections between FEM and PINN have been discussed in the literature (e.g., E and Yu, 2018; Kharazmi et al., 2021). Investigating the relationship between FEM and PILM may provide deeper insights into these methods.

## Abbreviations

BC: boundary condition; FEM, finite element method; GNSS, Global Navigation Satellite System; IC, initial condition; MAP, maximum a posteriori; ODE, ordinary differential equation; PDE, partial differential equation; PIGP, physics-informed Gaussian process; PILM, physics-informed linear model; PIML, physics-informed machine learning; PINN, physics-informed neural network.

## Acknowledgments

This study was supported by the Grant-in-Aid for Scientific Research (B) (Kakenhi No. 25K01084) from the Ministry of Education, Culture, Sports, Science and Technology (MEXT) of Japan. This study was also supported by MEXT, under its The Third Earthquake and Volcano Hazards Observation and Research Program (Earthquake and Volcano Hazard Reduction Research).

## Availability of data and materials

GNSS velocity data were obtained from Supplementary Information in Okazaki et al. (2021). The program codes for strain rate estimation are available at Okazaki (2025).

## Competing interests

The author declares that there is no known competing interest.

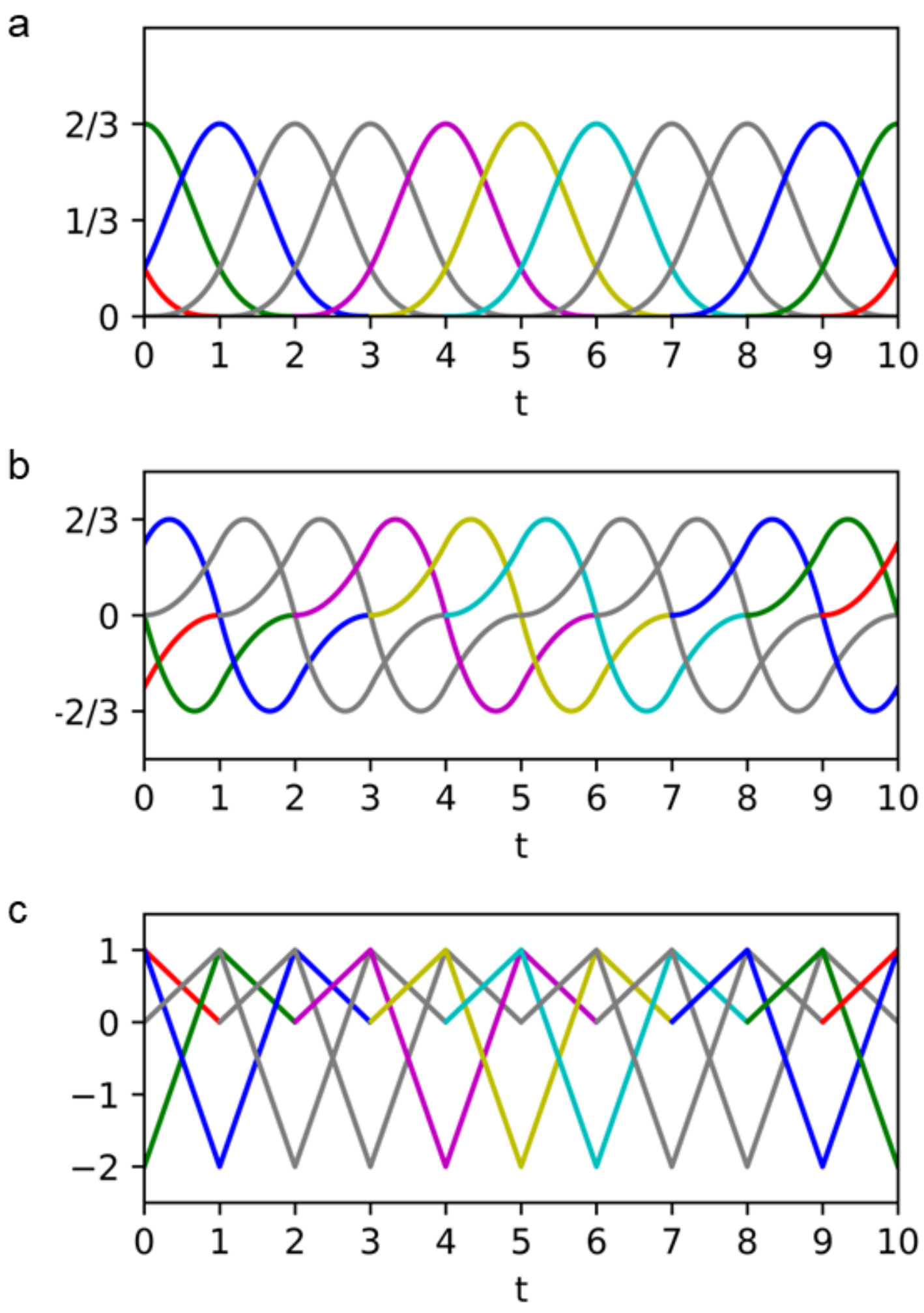

**Figure S1.** Shape of the (a) basis functions, (b) first-order derivatives, and (c) second-order derivatives, for $T = 10$ and $M = 13$ ($\Delta t = 1$). They are piecewise (a) cubic, (b) quadratic, and (c) linear functions.

$$
\mathbf{R}^{ab} = \begin{bmatrix}
a_{00} & a_{01} & a_{02} & K_3 & & & & & & & \\
a_{10} & a_{11} & a_{12} & K_2 & K_3 & & & & & \mathbf{0} & \\
a_{20} & a_{21} & a_{22} & K_1 & K_2 & K_3 & & & & & \\
K_{-3} & K_{-2} & K_{-1} & K_0 & K_1 & K_2 & K_3 & & & & \\
 & K_{-3} & K_{-2} & K_{-1} & K_0 & K_1 & K_2 & K_3 & & & \\
 & & \ddots & \ddots & \ddots & \ddots & \ddots & \ddots & \ddots & & \\
 & & & K_{-3} & K_{-2} & K_{-1} & K_0 & K_1 & K_2 & K_3 & \\
 & & & & K_{-3} & K_{-2} & K_{-1} & K_0 & K_1 & K_2 & K_3 \\
 & \mathbf{0} & & & & K_{-3} & K_{-2} & K_{-1} & b_{33} & b_{32} & b_{31} \\
 & & & & & & K_{-3} & K_{-2} & b_{23} & b_{22} & b_{31} \\
 & & & & & & & K_{-3} & b_{13} & b_{12} & b_{11}
\end{bmatrix}
$$

**Figure S2.** Form of the definite-integral matrix. The numerical values are presented in Table S1. The upper left and lower right 3×3 blocks result from truncation of the basis functions.

**Table S1.** Numerical values of the definite-integral matrix in Figure S2

| | | |
|---|---|---|
| $\mathbf{R}^{00}$ | $K$ | 151/315, 397/1680, 1/42, 1/5040, 397/1680, 1/42, 1/5040 |
| | $a$ | 1/252, 43/1680, 1/84, 43/1680, 151/630, 59/280, 1/84, 59/280, 599/1260 |
| | $b$ | 1/252, 43/1680, 1/84, 43/1680, 151/630, 59/280, 1/84, 59/280, 599/1260 |
| $\mathbf{R}^{11}$ | $K$ | 2/3, −1/8, −1/5, −1/120, −1/8, −1/5, −1/120 |
| | $a$ | 1/20, 7/120, −1/10, 7/120, 1/3, −11/60, −1/10, −11/60, 37/60 |
| | $b$ | 1/20, 7/120, −1/10, 7/120, 1/3, −11/60, −1/10, −11/60, 37/60 |
| $\mathbf{R}^{22}$ | $K$ | 8/3, −3/2, 0, 1/6, −3/2, 0, 1/6 |
| | $a$ | 1/3, −1/2, 0, −1/2, 4/3, −1, 0, −1, 7/3 |
| | $b$ | 1/3, −1/2, 0, −1/2, 4/3, −1, 0, −1, 7/3 |
| $\mathbf{R}^{01}$ | $K$ | 0, 49/144, 7/90, 1/720, −49/144, −7/90, −1/720 |
| | $a$ | −1/72, −1/80, 1/40, −71/720, −2/9, 29/120, −19/360, −127/360, −1/72 |
| | $b$ | 1/72, 1/80, −1/40, 71/720, 2/9, −29/120, 19/360, 127/360, 1/72 |
| $\mathbf{R}^{10}$ | $K$ | 0, −49/144, −7/90, −1/720, 49/144, 7/90, 1/720 |
| | $a$ | −1/72, −71/720, −19/360, −1/80, −2/9, −127/360, 1/40, 29/120, −1/72 |
| | $b$ | 1/72, 71/720, 19/360, 1/80, 2/9, 127/360, −1/40, −29/120, 1/72 |
| $\mathbf{R}^{02}$ | $K$ | −2/3, 1/8, 1/5, 1/120, 1/8, 1/5, 1/120 |
| | $a$ | 1/30, −7/120, 1/60, 11/40, −1/3, −3/20, 11/60, 11/60, −7/10 |
| | $b$ | 1/30, −7/120, 1/60, 11/40, −1/3, −3/20, 11/60, 11/60, −7/10 |
| $\mathbf{R}^{20}$ | $K$ | −2/3, 1/8, 1/5, 1/120, 1/8, 1/5, 1/120 |
| | $a$ | 1/30, 11/40, 11/60, −7/120, −1/3, 11/60, 1/60, −3/20, −7/10 |
| | $b$ | 1/30, 11/40, 11/60, −7/120, −1/3, 11/60, 1/60, −3/20, −7/10 |
| $\mathbf{R}^{12}$ | $K$ | 0, 19/24, −1/3, −1/24, −19/24, 1/3, 1/24 |
| | $a$ | −1/8, 5/24, −1/24, −5/24, 0, 7/12, 7/24, −7/12, −1/8 |
| | $b$ | 1/8, −5/24, 1/24, 5/24, 0, −7/12, −7/24, 7/12, 1/8 |
| $\mathbf{R}^{21}$ | $K$ | 0, −19/24, 1/3, 1/24, 19/24, −1/3, −1/24 |
| | $a$ | −1/8, −5/24, 7/24, 5/24, 0, −7/12, −1/24, 7/12, −1/8 |
| | $b$ | 1/8, 5/24, −7/24, −5/24, 0, 7/12, 1/24, −7/12, 1/8 |

$K = (K_0, K_1, K_2, K_3, K_{-1}, K_{-2}, K_{-3})$,
$a = (a_{00}, a_{01}, a_{02}, a_{10}, a_{11}, a_{12}, a_{20}, a_{21}, a_{22})$, $b = (b_{11}, b_{12}, b_{13}, b_{21}, b_{22}, b_{23}, b_{31}, b_{32}, b_{33})$.

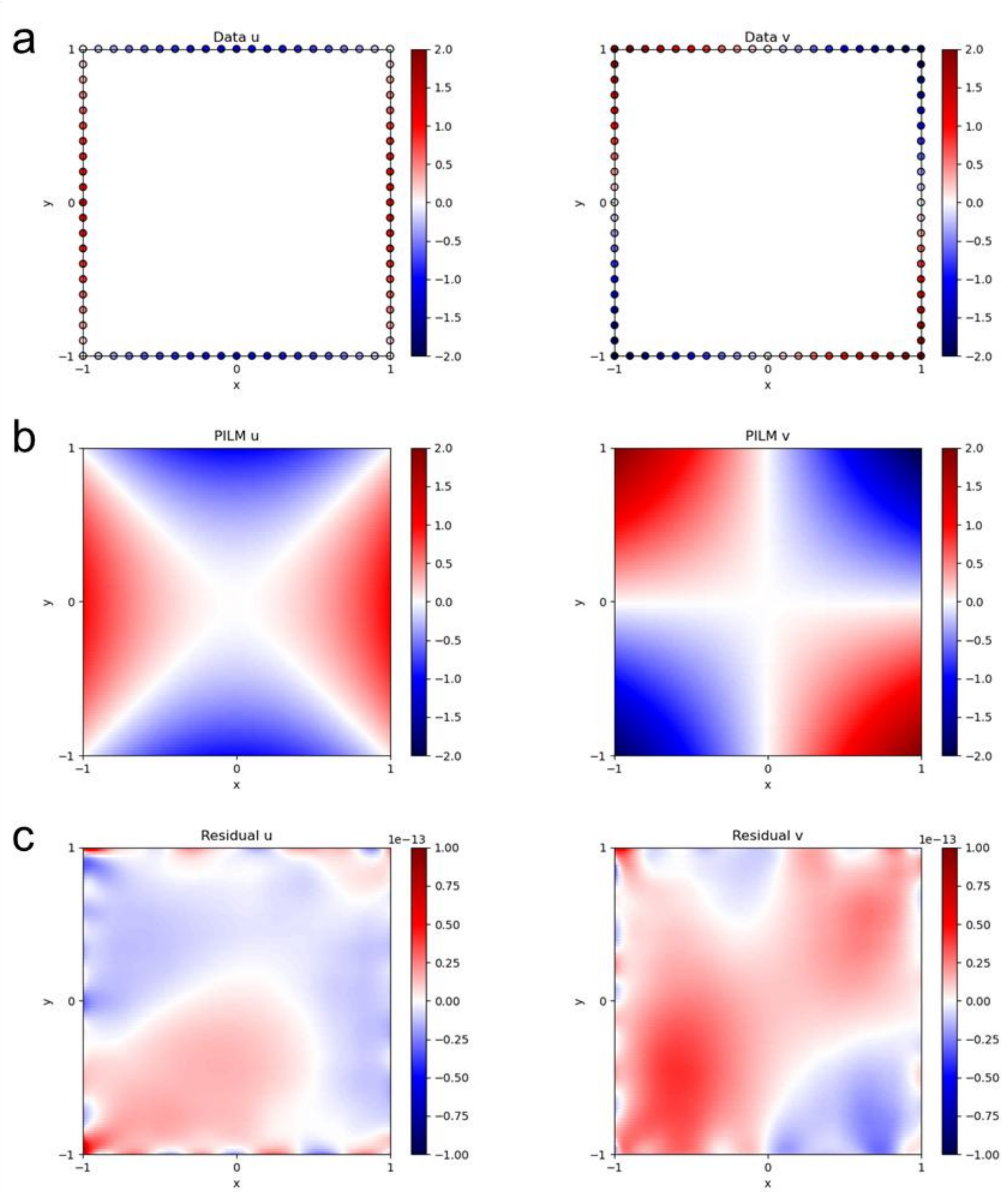

**Figure S3.** Verification of PILM for inplane elasticity. The left and right columns correspond to the two velocity components. (a) Input data. Values of an analytical solution $u = x^2 - y^2$ and $v = -2xy$ are provided at 80 points on the boundary. (b) PILM estimation using parameters $\nu = 0.5$ and $M = 23^2$. (c) Residual distribution of the PILM relative to the analytical solution. The absolute residual is less than $10^{-13}$ across the entire domain.

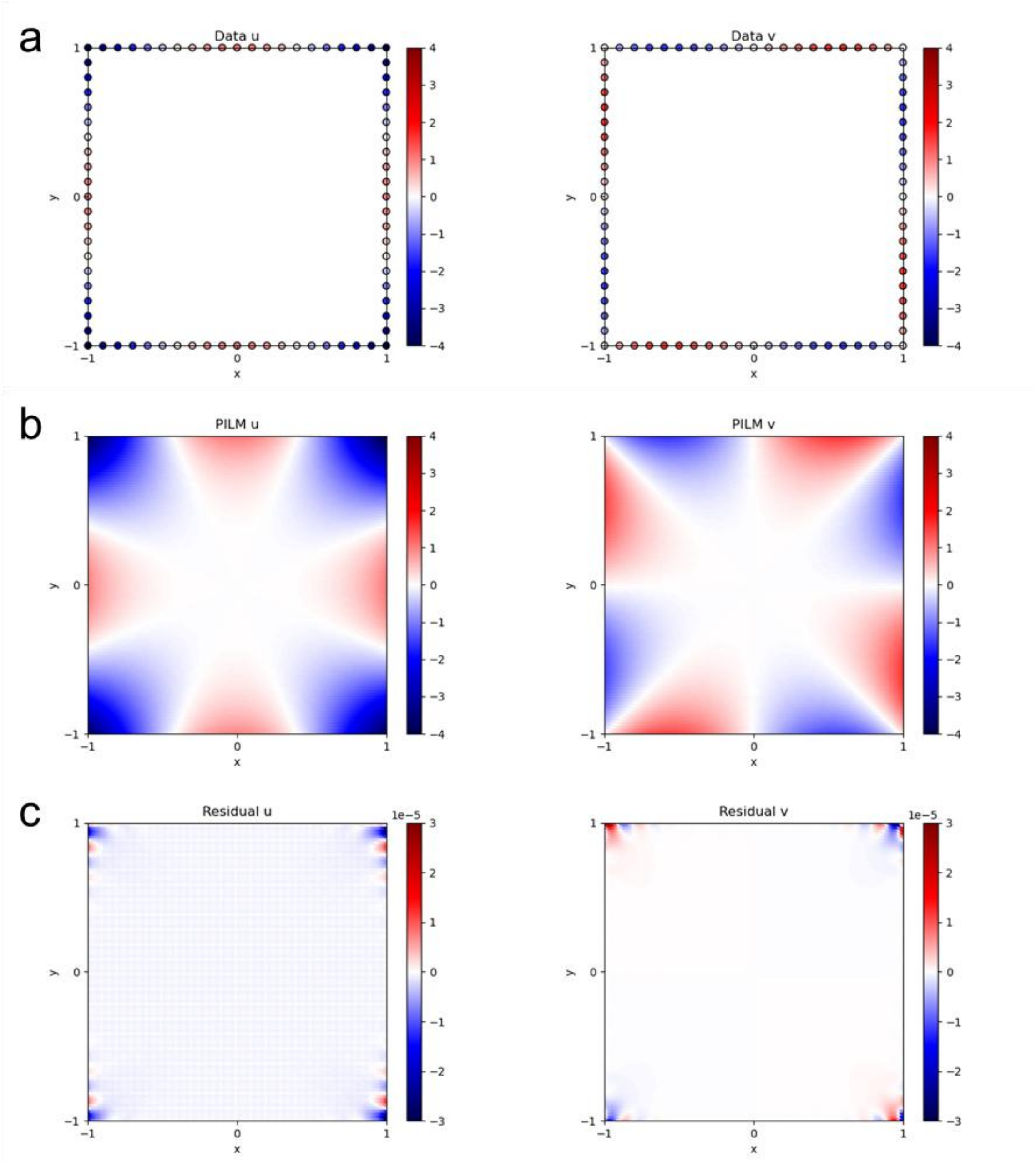

**Figure S4.** Same as Figure S3, but for an analytical solution $u = x^4 - 6x^2y^2 + y^4$ and $v = -4xy(x^2 - y^2)$, with PILM parameters of $\nu = 0$ and $M = 43^2$. The absolute residual is less than $3\times10^{-5}$ across the entire domain.